# A Self-Adaptive Synthetic Over-Sampling Technique for Imbalanced Classification

Xiaowei Gu[1,2,*], Plamen P Angelov[1,2,3*] and Eduardo Almeida Soares[1,2]

[1] School of Computing and Communications, Lancaster University, Lancaster, UK, LA1 4WA

[2] Lancaster Intelligent, Robotic and Autonomous Systems Centre, Lancaster University, Lancaster, UK

[3] Honorary Professor at Technical University, Sofia, 1000, Bulgaria

[*] The first two authors contribute equally.

Email: x.gu3@lancaster.ac.uk; p.angelov@lancaster.ac.uk; e.almeidasoares@lancaster.ac.uk

**Abstract:** Traditionally, in supervised machine learning, (a significant) part of the available data (usually 50% to 80%) is used for training and the rest – for validation. In many problems, however, the data is highly imbalanced in regard to different classes or does not have good coverage of the feasible data space which, in turn, creates problems in validation and usage phase. In this paper, we propose a technique for synthesising feasible and likely data to help balance the classes as well as to boost the performance in terms of confusion matrix as well as overall. The idea, in a nutshell, is to synthesise data samples in close vicinity to the actual data samples specifically for the less represented (minority) classes. This has also implications to the so-called fairness [1] of machine learning. In this paper, we propose a specific method for synthesising data in a way to balance the classes and boost the performance, especially of the minority classes. It is generic and can be applied to different base algorithms, e.g. support vector machine, k-nearest neighbour, deep networks, rule-based classifiers, decision trees, etc. The results demonstrated that: *i)* a significantly more balanced (and fair) classification results can be achieved; *ii)* that the overall performance as well as the performance per class measured by confusion matrix can be boosted. In addition, this approach can be very valuable for the cases when the number of actual available labelled data is small which itself is one of the problems of the contemporary machine learning.

## 1. Introduction

In machine learning, classification is to learn a predictive model from training data that can perform accurate prediction on the categories of previously unseen data. Most of standard classification approaches are designed for larger-scale and balanced data sets with the goal of maximizing overall classification accuracy [2]. For example, let us consider an extreme case, if a data set consists of five samples of class 1 and 995 samples of class 2, a classifier can achieve 99.5% accuracy even if it classifies all data samples as class 1. As a result, classifiers learned from imbalanced data sets tend to ignore the minority class because the minority class samples are outnumbered by the majority class samples and they play a much weaker role in the overall performance evaluation.

On the other hand, the class imbalance problem often occurs in real-world applications, e.g., financial fraud detection [3], medical diagnosis [4] and mechanical fault detection [5], where minorities (rare samples) are of greater interest. In such application scenarios, the primary goal for classification algorithms is to identify the rare samples as accurately as possible because the costs of misclassifying the minorities can be very high. Although traditional classification algorithms usually demonstrate good performance on standard classification tasks, they usually straggle with imbalanced problems [6]. In recent years, imbalanced learning have received much more attentions and is now a hotly studied problem in the fields of machine learning and pattern recognition [7]. Till now, many successful methods have been proposed for tackling the class imbalance problem [7], [8], which can be categorizes into three major groups [6], namely, *1)* data sampling, *2)* cost-sensitive learning and *3)* algorithmic modification. A review of the state-of-the-art will be given in the next section.

In this paper, the focus of our study is synthetic data sampling. More specifically, we propose a novel self-adaptive synthetic over-sampling (SASYNO) approach to tackle the class imbalance problem. The common practice of popular data sampling approaches is to randomly select minority class samples and create linear interpolations between them and their neighbours for artificial data synthesis. However, this strategy is not necessarily going to expand the knowledge base but are more likely to create overlaps between the expanded minority class and the original majority class, especially when data structure is highly complex. In contrast, the key idea of the proposed approach is to select out neighbouring minority class samples based on their mutual distances and create both interpolations and extrapolations around neighbouring samples for synthetic data generation. SASYNO firstly identifies a population of pairwise neighbouring samples from minority class. Then, SASYNO imposes Gaussian disturbance on these identified neighbouring samples to create extrapolations, and, finally, generates synthetic samples by creating linear interpolations between these extrapolations. Comparing with standard data sampling approaches, the uniqueness of SASYNO comes from the following two aspects:

*1)* The proposed approach selects out the most proper candidates from minority class samples and uses them for data synthesis only. This allows SASYNO to precisely expand the minority class avoiding possible overlaps with the majority class.

*2)* The proposed approach employs Gaussian disturbance to create extrapolations from existing data samples for synthetic data generation, which gives SASYNO an extra degree of freedom for expanding the knowledge base.

The remainder of this paper is organized as follows. Section 2 provides a review of related works. The algorithmic details of SASYNO are given in Section 3. Section 4 presents numerical examples as the proof of concept with detailed analysis and discussions. This paper is concluded by Section 5.

## 2. Related Work

As mentioned in the previous section, popular approaches for imbalance learning generally can be categorized into three major types: *1)* data sampling, *2)* cost-sensitive learning and *3)* algorithmic modification.

- Data sampling approaches [7]–[9] rebalance the data sets by sampling, which is achieved by over-sampling the minority class [7], under-sampling the majority class [10] or a hybrid of both [11].
- Cost-sensitive learning approaches [4], [12] incorporate the costs of misclassifying minority class samples into function minimization.
- Algorithmic modification approaches [13], [14] are the modifications of commonly-used machine learning algorithms to achieve better performance with imbalanced data set.

Currently, data sampling approaches are the dominant solutions to address the class imbalance problem because they are more generic and can be employed by standard classification methods [15].

Random over-sampling and down-sampling are the two basic and easy-to-use approaches for balancing data through randomly duplicating or removing samples from the minority or majority classes. However, in many class imbalance problems, minor class samples are much rare compared with the majority class samples. Down-sampling of the majority class is not advisable since it will cause a significant loss of information. Therefore, over-sampling techniques are more popular and intensively studied [16].

The most successful advanced over-sampling approaches are SMOTE (synthetic minority over-sampling technique) [8], ADASYN (adaptive synthetic sampling approach) [7] and MWMOTE (majority weighted minority over-sampling technique) [9]. SMOTE [8] tackles the class imbalance problem by creating linear interpolations between randomly selected minority class samples and their neighbours of the same class. ADASYN [7] uses a very similar strategy as SMOTE, but it prioritizes samples near decision boundaries and focuses on these hard-to-learn minority class samples by assigning weights calculated per sample as the ratio of neighbours belonging to the majority class. MWMOTE [9] firstly identifies the minority class samples at the decision boundaries and assigns them weights based on their distances to neighbouring majority class samples. Then, MWMOTE clusters these minority class samples for generating the synthetic samples.

Other popular over-sampling approaches include Borderline-SMOTE (BLSMOTE) by Han et al. [17], Safe-Level-SMOTE (SLSMOTE) by Bunkhumpornpat et al. [18], RACOG (rapidly converging Gibbs algorithm) by Das et al. [16], MDO (Mahalanobis distance-based over-sampling algorithm) by Adbi and Hashemi [19], A-SUWO (adaptive semi-unsupervised weighted oversampling algorithm) by Nekooeimehr and Lai-Yuen [20] and SMOM (*k*-nearest neighbours-based synthetic minority oversampling algorithm) by Zhu et al. [21], etc. However, due to the limited space of this paper, it is impossible to cover all the data sampling approaches in the literature, interested readers may refer to [6], [22], [23] for more details.

## 3. Proposed Method

In this section, details of the proposed SASYNO are presented. First of all, let $\{x\}_N = \{x_1, x_2, \ldots, x_N\}$ ($x_i = [x_{i,1}, x_{i,2}, \ldots, x_{i,M}]^T \in \mathbf{R}^M$) be a two-class data set in a real data space, $\mathbf{R}^M$, where $M$ is the dimensionality; the subscript $i$ denotes the index of $x_i$. The data set consists of two classes, namely, "Class 0" and "Class 1". According to class labels, $\{x\}_N$ can be divided into two sets, $\{x\}_{N^0}^0$ and $\{x\}_{N^1}^1$, where $N^0$ and $N^1$ are the respective numbers of data samples of the two classes and the superscripts "0" and "1" indicate the class labels. In this paper, we assume that class 0 is the minority class, and class 1 is the majority one, namely, $N^0 < N^1$.

The algorithmic procedure of SASYNO is described as follows. By default, SASYNO is used for generating synthetic minority class data samples to balance the minority and majority classes. Nonetheless, SASYNO is, in principle, a generic approach for data augmentation and can be used for generating any amount of synthetic data samples of any classes.

**Stage 1. Identifying pairwise neighbouring samples**

In this stage, we identify neighbouring data samples based on the ensemble properties and mutual distribution of the minority class samples, $\{x\}_{N^0}^0$. In order to define the concept of closeness directly from the observed data, we employ the following objective quantifier of the data pattern [24]:

$$\gamma = \frac{1}{P_\mu} \sum_{x_i^0, x_j^0 \in \{x\}_{N^0}^0;\, \|x_i^0 - x_j^0\| \le \mu; i \ne j} \|x_i^0 - x_j^0\| \tag{1}$$

where $\|x_i^0 - x_j^0\|$ denotes the Euclidean distance between $x_i^0$ and $x_j^0$, $\|x_i^0 - x_j^0\| = \sqrt{(x_i^0 - x_j^0)^T (x_i^0 - x_j^0)}$; $\mu$ is the average distance between any pair of minority class samples, $\mu = \frac{2}{N^0(N^0-1)} \sum_{i=1}^{N^0-1} \sum_{j=i+1}^{N^0} \|x_i^0 - x_j^0\|$; $\gamma$ is the average distance between any pair of minority class samples between which the distance is less than $\mu$; $P_\mu$ is the number of such pairs. The quantifier $\gamma$ provides an estimation of average distance between any two data samples that are considered as spatially neighbouring. Based on this quantifier. Note that $\gamma$ is directly derived from data without making any prior assumptions on generation model with parameters.

Based on the objectively derived quantifier, $\gamma$, we can identify a collection of pair-wise neighbouring samples from the minority class samples, denoted by, $\mathbf{P}$ using the following condition:

$$IF\ (\|x_i^0 - x_j^0\| \le \gamma) \quad THEN\ \left(\mathbf{P} \leftarrow \mathbf{P} \cup \{(p_k, q_k) = (x_i^0, x_j^0)\};\ k \leftarrow k+1\right) \tag{2}$$

where $x_i^0, x_j^0 \in \{x\}_{N^0}^0$ and $i \ne j$. The identified pair-wise neighbouring samples will be used for generating synthetic samples in the next stages.

The rationale behind the pair-wise neighbouring samples identification is to identify the subspaces which are occupied by the minority class samples only. Synthetic samples generated around these subspaces are highly unlikely to be overlapped with the major class samples. Thus, the quality of the synthetic samples is guaranteed.

**Stage 2. Creating explorations by Gaussian disturbance**

In this stage, the algorithm randomly selects a pair of neighbouring samples from the collection $\mathbf{P}$ denoted by $(p_k, q_k)^* \in \mathbf{P}$ ($k \leftarrow 1$) and apply Gaussian disturbance to create extrapolations in the data space:

$$(\hat{p}_k, \hat{q}_k)^* = (p_k + g_p, q_k + g_q)_k^* \tag{3}$$

where $g_p = [g_{p,1}, g_{p,2}, \ldots, g_{p,M}]^T$ and $G_q = [g_{q,1}, g_{q,2}, \ldots, g_{q,M}]^T$ are two $M$ dimensional randomly generated vectors following the Gaussian distributions, $g_{p,l}, g_{q,l} \sim \aleph(0, \sigma_l^2)$ ($l = 1, 2, \ldots, M$); the standard deviation, $\sigma_l$ of the Gaussian distribution $\aleph(0, \sigma_l^2)$ is defined per attribute in a similar way to $\gamma$ (equation (1)) as follows:

$$\sigma_l = \frac{1}{P_{\mu_l}} \sum_{x_i^0, x_j^0 \in \{x\}_{N^0}^0;\, |x_{i,l}^0 - x_{j,l}^0| \le \mu_l; i \ne j} |x_{i,l}^0 - x_{j,l}^0| \tag{4}$$

where $|\cdot|$ denotes the absolute value; $\mu_l = \frac{2}{N^0(N^0-1)} \sum_{i=1}^{N^0-1} \sum_{j=i+1}^{N^0} |x_{i,l}^0 - x_{j,l}^0|$ is the average distance between any two data samples belonging to $\{x\}_{N^0}^0$ at the $l$th dimension of $\mathbf{R}^M$. By applying Gaussian disturbance to neighbouring sample pairs, the subspaces occupied by the minority class samples are extended in an exploratory

way, which gives the proposed algorithm an extra degree of freedom to extrapolate new knowledge from the empirically observed data.

**Stage 3. Creating interpolations for synthetic data generation**

In this stage, the algorithm generates a synthetic sample by creating random interpolation between $\hat{\boldsymbol{p}}_k$ and $\hat{\boldsymbol{q}}_k$:

$$\boldsymbol{s}_k = \boldsymbol{r}_k^T \hat{\boldsymbol{p}}_k + (1 - \boldsymbol{r}_k)^T \hat{\boldsymbol{q}}_k \qquad (5)$$

where $\boldsymbol{r}_k = [r_{k,1}, r_{k,2}, \ldots, r_{k,M}]^T$ is a $M$ dimensional random vector, each element, $r_{k,l}$ ($l = 1, 2, \ldots, M$) follows the uniform distribution with the value range of $[0,1]$. Then, the algorithm goes back to Stage 2 to create the next synthetic sample ($k \leftarrow k + 1$).

To balance the data set, $N_s^0$ ($N_s^0 = N^1 - N^0$) extra minority class samples are needed, therefore, the same process will be repeated for $N_s^0$ times. Once $\{\boldsymbol{s}\}_{N_s^0} = \{\boldsymbol{s}_1, \boldsymbol{s}_2, \ldots, \boldsymbol{s}_{N_s^0}\}$ are generated from the selected neighbouring sample pairs, they are merged into $\{\boldsymbol{x}\}_{N^0}^0$: $\{\boldsymbol{x}\}_{N^0}^0 \leftarrow \{\boldsymbol{x}\}_{N^0}^0 \cup \{\boldsymbol{s}\}_{N_s^0}$, and the minority and majority classes are balanced.

An illustration of the syntenic data generation process introduced by this paper is given in Fig. 1. The yellow and blue ellipsoids surrounding $\boldsymbol{p}_k$ and $\boldsymbol{q}_k$ are the areas that $\hat{\boldsymbol{p}}_k$ and $\hat{\boldsymbol{q}}_k$ are highly likely to appear in the data space after Gaussian disturbance. The radii of the yellow and blue ellipsoids surrounding $\boldsymbol{p}_k$ and $\boldsymbol{q}_k$ are $2\sigma$ and $3\sigma$, respectively. According to the "68–95–99.7" rule, the probability for $\hat{\boldsymbol{p}}_k$ and $\hat{\boldsymbol{q}}_k$ to appear within the respective yellow ellipsoids is 90.3% (95% × 95%) and the probability for $\hat{\boldsymbol{p}}_k$ and $\hat{\boldsymbol{q}}_k$ to appear within the respective blue ellipsoids is 99.4% (99.7% × 99.7%). The yellow and blue capsules are the areas that $\boldsymbol{s}_k$ (generated from $\boldsymbol{p}_k$ and $\boldsymbol{q}_k$) would appear with a very large chance. The probability for $\boldsymbol{s}_k$ to appear within the yellow capsule is 81.5% (90.3% × 90.3%) and the probability is 98.8% (99.4% × 99.4%) for $\boldsymbol{s}_k$ to appear in the blue capsule.

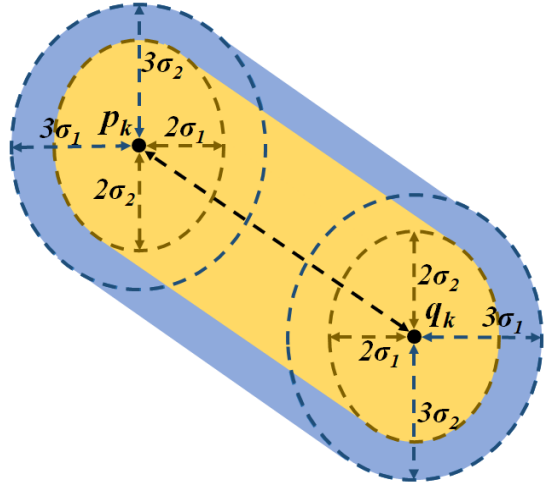

Fig. 1. Illustration of generating synthetic data from $\boldsymbol{p}_k$ and $\boldsymbol{q}_k$.

The main procedure of SASYNO is summarized in the form of pseudo-code as follows.

| Input: $\{\boldsymbol{x}\}_N$ |
| --- |
| *i.* Calculate the amount of synthetic data samples needed to be generated: $N_s^0 = N^1 - N^0$; |
| *ii.* Calculate the quantifier of the data pattern, γ by equation (1); |
| *iii.* Identify pair-wise neighbouring samples, **P** from $\{\boldsymbol{x}\}_{N^0}^0$ by equation (2); |
| *iv.* For $k = 1$ to $N_s^0$ do |
|     1. Randomly select a pair of neighbouring samples, $(\boldsymbol{p}_k, \boldsymbol{q}_k)^*$ from **P**; |
|     2. Apply Gaussian disturbance to $(\boldsymbol{p}_k, \boldsymbol{q}_k)^*$ by equation (3) and obtain $(\hat{\boldsymbol{p}}_k, \hat{\boldsymbol{q}}_k)^*$; |
|     3. Create random interpolation between $(\hat{\boldsymbol{p}}_k, \hat{\boldsymbol{q}}_k)^*$ and obtain $\boldsymbol{s}_k$; |

| | |
|---|---|
| *v.* End for | |
| *vi.* $\{x\}_{N^0}^0 \leftarrow \{x\}_{N^0}^0 \cup \{s\}_{N_s^0}$ | |
| Output: $\{x\}_N$ | |

## 4. Numerical Examples and Discussion

In this section, we evaluate the performance of SASYNO based on a variety of real-world data sets.

### 4.1. Numerical examples on imbalanced data sets

Since the imbalanced binary classification problem is the primary focus of the recent researches, SASYNO is firstly tested on five popular data sets as summarized in Table 1. In this paper, for all binary classification problems, the minority and majority classes are re-denoted by "Class 0" and "Class 1", respectively.

Table 1. Details of binary classification data sets used for performance evaluation

| Dataset | | # Samples, $N_s$ | # Minority, $N_0$ (Class 0) | # Majority, $N_1$ (Class 1) | #Attributes, $M$ |
|---|---|---|---|---|---|
| Wilt (WI)[1] | Training Set | 4339 | 74 | 4265 | 5 + 1 label |
| | Testing Set | 500 | 187 | 313 | |
| Spambase (SB)[2] | | 4601 | 1813 | 2788 | 57 + 1 label |
| German Credit (GC)[3] | | 1000 | 300 | 7000 | 24 + 1 label |
| Mammograph (MG)[4] | | 11183 | 260 | 10923 | 6 + 1 label |
| Occupancy Detection (OD)[5,6] | Training Set | 8143 | 1729 | 6414 | 5 + 1 label |
| | Testing Set 1 | 2665 | 972 | 1693 | |
| | Testing Set 2 | 9752 | 2049 | 7703 | |

[1]Available from http://archive.ics.uci.edu/ml/datasets/wilt
[2]Available from https://archive.ics.uci.edu/ml/datasets/Spambase
[3]Available from https://archive.ics.uci.edu/ml/datasets/Statlog+(German+Credit+Data)
[4]Available from http://odds.cs.stonybrook.edu/mammography-dataset/
[5]Available from https://archive.ics.uci.edu/ml/datasets/Occupancy+Detection+
[6]Time stamps have been removed in advance

The following six classification approaches are involved as the base classifiers:

*1)* Self-organizing neuro-fuzzy inference system (SONFIS) [25];

*2)* Support vector machine classifier (SVM) [26];

*3)* k-nearest neighbour classifier (KNN) [27];

*4)* decision tree classifier (DT) [28];

*5)* Random forest classifier (RF) [29], and;

*6)* Multilayer perceptron (MLP).

Here SONFIS uses Euclidean distance as the distance measure, and the level of granularity is set as 12; SVM uses Gaussian kernel; $k$ is equal to 1 for KNN; RF uses an ensemble of 100 classification trees; MLP is composed of one input layer, two hidden layers and one output layer, each hidden layers has 20 neurons.

The quality of the synthetic data generated by the proposed approach is also compared with the synthetic data generated by the state-of-the-art approaches as follows.

*1)* ADASYN [7];

*2)* SMOTE [8];

*3)* BLSMOTE [17];

*4)* SLSMOTE [18], and;

*5)* Random down-sampling (RDS).

In this paper, for ADASYN, SMOTE, BLSMOTE and SLSMOTE algorithms, the number of nearest neighbours is set as $k=5$. During the experiments, all the involved over-sampling approaches (SASYNO, ADASYN, SMOTE, BLSMOTE and SLSMOTE) generate $N_1 - N_0$ new synthetic samples from the minority class samples to balance the minority and majority classes. RDS randomly remove $N_1 - N_0$ majority class samples to achieve the same purpose.

For imbalanced classification problems, the commonly-used performance criterion, namely, overall accuracy is insufficient for evaluation. Thus, we further involve a set of assessment metrics related to receiver operating characteristics graph as follows [7], [8].

*1)* Sensitivity (*SN*). *SN* is the true positive ratio, also known as recall: $SN = \frac{TP}{TP+FN}$;

*2)* Specificity (*SP*). *SP* is the true negative ratio measured defined as: $SP = \frac{TN}{TN+FP}$;

*3)* F-Measure (*FM*): $FM = \frac{2TP}{2TP+FN+FP}$;

*4)* G-mean (*GM*): $GM = \sqrt{\frac{TP}{TP+FN} \cdot \frac{TN}{TN+FP}}$;

*5)* Overall accuracy (*Acc*): $Acc = \frac{TP+TN}{TP+FN+FP+TN}$;

The definitions of *TP*, *TN*, *FP*, *FN* are given by Fig. 2, where $K_0$ and $K_1$ represent the numbers of minority and majority class samples in the prediction results, respectively.

|  | Hypothesis Output | |
|---|---|---|
| True Class | $K_0$ | $K_1$ |
| $N_0$ | TP (True Positives) | FN (False Negatives) |
| $N_1$ | FP (True Negatives) | TN (True Negatives) |

Fig. 2. Confusion matrix

In the first numerical example, we use SONFIS as the base classifier to evaluate the performance of the proposed SASYNO and compare with the alternative data sampling approaches on WI, SB, GC, MG and OD data sets. For SB, GC and MG data sets, 80% data samples are randomly selected out as the training set, and the remaining are used for testing. We keep the original splitting for WI and OD data sets, but combine the testing set 1 and 2 of the OD data set as one. The obtained results after 10 Monte Carlo experiments in terms of the five performance measures given above are tabulated in Table 1. The performance of SONFIS on original data (denoted as ORIG) is also reported in Table 1 as the baseline. The respective ranks per performance measure per data set are reported in the same table (in italic) for visual clarity. The average ranks per performance measure across the five data sets are also reported at the end of this table. For better illustration, we visualize the obtained synthetic data samples by SASYNO together with the original data samples using the t-SNE technique [30] in Fig. 3, where one can clearly see that SASYNO significantly expands the minority class and effectively avoids overlaps with the majority class.

Table 1. Performance comparison between data sampling approaches using SONFIS as base learner

| Dataset | Algorithm | *SN* | *SP* | *GM* | *FM* | *Acc* |
|---|---|---|---|---|---|---|
| WI | SASYNO | 0.7319 | 0.9183 | 0.8198 | 0.7989 | 0.8344 |
|  |  | *6* | *1* | *6* | *1* | *2* |
|  | ADASYN | 0.7887 | 0.8765 | 0.8314 | 0.7914 | 0.8434 |
|  |  | *5* | *2* | *3* | *2* | *1* |
|  | SMOTE | 0.9284 | 0.7517 | 0.8354 | 0.6129 | 0.7842 |
|  |  | *2* | *6* | *2* | *6* | *5* |
| | BLSMOTE | 0.8409 *4* | 0.8142 *3* | 0.8274 *4* | 0.7305 *4* | 0.8218 *4* |
| | SLSMOTE | 0.8833 *3* | 0.8085 *4* | 0.8451 *1* | 0.7306 *3* | 0.8282 *3* |
| | RDS | 0.5445 *7* | 0.8024 *5* | 0.6608 *7* | 0.6275 *5* | 0.6696 *7* |
| | ORIG | 0.9333 *1* | 0.7247 *7* | 0.8224 *5* | 0.5344 *7* | 0.7560 *6* |
| SB | SASYNO | 0.7016 *4* | 0.8616 *2* | 0.7773 *2* | 0.7492 *2* | 0.7895 *2* |
| | ADASYN | 0.6733 *6* | 0.8694 *1* | 0.7650 *6* | 0.7417 *4* | 0.7753 *6* |
| | SMOTE | 0.7530 *1* | 0.8470 *6* | 0.7985 *1* | 0.7585 *1* | 0.8096 *1* |
| | BLSMOTE | 0.6857 *5* | 0.8546 *4* | 0.7653 *5* | 0.7372 *5* | 0.7775 *5* |
| | SLSMOTE | 0.7083 *3* | 0.8501 *5* | 0.7758 *3* | 0.7431 *3* | 0.7886 *3* |
| | RDS | 0.6620 *7* | 0.8549 *3* | 0.7522 *7* | 0.7267 *6* | 0.7632 *7* |
| | ORIG | 0.7141 *2* | 0.8284 *7* | 0.7689 *4* | 0.7260 *7* | 0.7818 *4* |
| GC | SASYNO | 0.3880 *6* | 0.7542 *3* | 0.5397 *6* | 0.4560 *3* | 0.5910 *6* |
| | ADASYN | 0.4052 *5* | 0.7596 *2* | 0.5533 *3* | 0.4632 *2* | 0.6120 *5* |
| | SMOTE | 0.4474 *1* | 0.7447 *6* | 0.5747 *1* | 0.4244 *6* | 0.6590 *1* |
| | BLSMOTE | 0.4102 *4* | 0.7504 *5* | 0.5527 *4* | 0.4457 *5* | 0.6220 *4* |
| | SLSMOTE | 0.4320 *2* | 0.7647 *1* | 0.5735 *2* | 0.4717 *1* | 0.6410 *2* |
| | RDS | 0.3741 *7* | 0.7512 *4* | 0.5289 *7* | 0.4528 *4* | 0.5720 *7* |
| | ORIG | 0.4132 *3* | 0.7364 *7* | 0.5501 *5* | 0.4082 *7* | 0.6385 *3* |
| MG | SASYNO | 0.3061 *3* | 0.9947 *2* | 0.5512 *3* | 0.4399 *2* | 0.9532 *2* |
| | ADASYN | 0.3366 *2* | 0.9945 *3* | 0.5777 *1* | 0.4685 *1* | 0.9586 *1* |
| | SMOTE | 0.4092 *1* | 0.9888 *6* | 0.5658 *2* | 0.3964 *3* | 0.8656 *3* |
| | BLSMOTE | 0.2788 *4* | 0.9922 *4* | 0.4797 *4* | 0.3608 *4* | 0.8518 *4* |
| | SLSMOTE | 0.1499 *6* | 0.9921 *5* | 0.3321 *5* | 0.2016 *5* | 0.7463 *6* |
| | RDS | 0.07290 *7* | 0.9958 *1* | 0.2566 *7* | 0.1289 *7* | 0.6353 *7* |
| | ORIG | 0.1709 *5* | 0.9868 *7* | 0.3310 *6* | 0.1836 *6* | 0.7488 *5* |
| OD | SASYNO | 0.3061 *3* | 0.9947 *2* | 0.5512 *3* | 0.4399 *2* | 0.9532 *2* |
| | ADASYN | 0.3366 *2* | 0.9945 *3* | 0.5777 *1* | 0.4685 *1* | 0.9586 *1* |
| | SMOTE | 0.4092 *1* | 0.9888 *6* | 0.5658 *2* | 0.3964 *3* | 0.8656 *3* |
| | BLSMOTE | 0.2788 | 0.9922 | 0.4797 | 0.3608 | 0.8518 |

|  |  | 4 | 4 | 4 | 4 | 4 |
|---|---|---|---|---|---|---|
|  | SLSMOTE | 0.1499 | 0.9921 | 0.3321 | 0.2016 | 0.7463 |
|  |  | 6 | 5 | 5 | 5 | 6 |
|  | RDS | 0.07290 | 0.9958 | 0.2566 | 0.1289 | 0.6353 |
|  |  | 7 | 1 | 7 | 7 | 7 |
|  | ORIG | 0.1709 | 0.9868 | 0.3310 | 0.1836 | 0.7488 |
|  |  | 5 | 7 | 6 | 6 | 5 |
| Average Rank | SASYNO | 4.6 | 1.8 | 4 | 2 | 3 |
|  | ADASYN | 4.8 | 2.4 | 3.8 | 3 | 3.8 |
|  | SMOTE | 1.6 | 6.2 | 2 | 4.2 | 3 |
|  | BLSMOTE | 3.8 | 4.2 | 3.8 | 4.2 | 3.8 |
|  | SLSMOTE | 3.8 | 3.4 | 3.2 | 3.2 | 3.6 |
|  | RDS | 7 | 3.2 | 7 | 5.8 | 7 |
|  | ORIG | 2.4 | 6.8 | 4.2 | 5.6 | 3.8 |

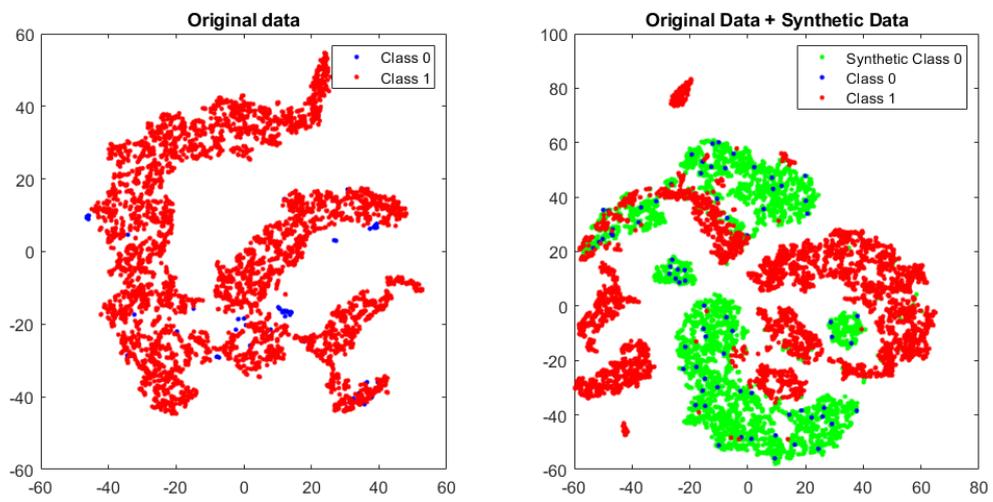

(a) WI data set

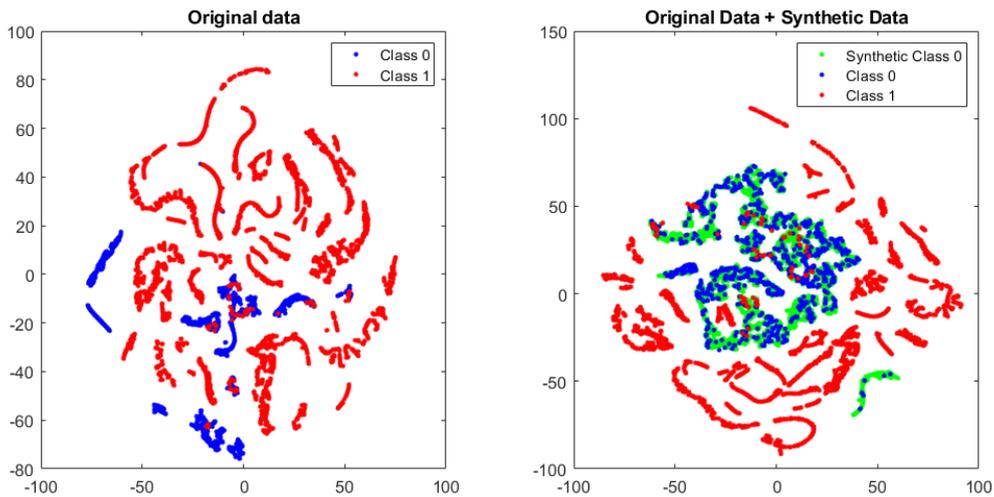

(b) OD data set

Fig. 3. Data visualization with t-SNE

In the second example, we use SVM and KNN as the base classifiers and repeat the experiments under the same protocol as the previous example. Numerical results are given in Tables 2 and 3, respectively. Additionally, we calculate the average ranks of the involved data sampling approaches across the three numerical examples and report it in Table 4.

Table 2. Performance comparison between data sampling approaches using SVM as base learner

| Dataset | Algorithm | SN | SP | GM | FM | Acc |
|---|---|---|---|---|---|---|
| WI | SASYNO | 0.3755 | 1.0000 | 0.6128 | 0.5460 | 0.3780 |
|  |  | 2 | 1 | 2 | 1 | 2 |
|  | ADASYN | 1.0000 | 0.6273 | 0.7920 | 0.0106 | 0.6280 |
|  |  | 1 | 2 | 1 | 3 | 1 |
|  | SMOTE | 1.0000 | 0.6273 | 0.7920 | 0.0106 | 0.6280 |
|  |  | 1 | 2 | 1 | 3 | 1 |
|  | BLSMOTE | 1.0000 | 0.6273 | 0.7920 | 0.0106 | 0.6280 |
|  |  | 1 | 2 | 1 | 3 | 1 |
|  | SLSMOTE | 1.0000 | 0.6273 | 0.7920 | 0.0106 | 0.6280 |
|  |  | 1 | 2 | 1 | 3 | 1 |
|  | RDS | 0.3745 | 0.3000 | 0.1838 | 0.5449 | 0.3752 |
|  |  | 3 | 3 | 3 | 2 | 3 |
|  | ORIG | 1.0000 | 0.6273 | 0.7920 | 0.0106 | 0.6280 |
|  |  | 1 | 2 | 1 | 3 | 1 |
| SB | SASYNO | 0.9271 | 0.7388 | 0.8206 | 0.5512 | 0.7377 |
|  |  | 7 | 1 | 6 | 1 | 1 |
|  | ADASYN | 0.9560 | 0.6996 | 0.8177 | 0.4987 | 0.7349 |
|  |  | 6 | 2 | 7 | 2 | 2 |
|  | SMOTE | 0.9948 | 0.6932 | 0.8303 | 0.4743 | 0.7301 |
|  |  | 2 | 5 | 1 | 6 | 5 |
|  | BLSMOTE | 0.9868 | 0.6932 | 0.8270 | 0.4748 | 0.7297 |
|  |  | 4 | 6 | 3 | 5 | 6 |
|  | SLSMOTE | 0.9915 | 0.6955 | 0.8303 | 0.4831 | 0.7328 |
|  |  | 3 | 4 | 2 | 4 | 3 |
|  | RDS | 0.9831 | 0.6958 | 0.8270 | 0.4845 | 0.7326 |
|  |  | 5 | 3 | 4 | 3 | 4 |
|  | ORIG | 0.9953 | 0.6863 | 0.8264 | 0.4477 | 0.7214 |
|  |  | 1 | 7 | 5 | 7 | 7 |
| GC | SASYNO | 0.3135 | 0.9333 | 0.5396 | 0.4764 | 0.3235 |
|  |  | 1 | 1 | 1 | 1 | 3 |
|  | ADASYN | 0.0000 | 0.6905 | 0.0000 | 0.0000 | 0.6905 |
|  |  | 3 | 3 | 3 | 3 | 1 |
|  | SMOTE | 0.0000 | 0.6905 | 0.0000 | 0.0000 | 0.6905 |
|  |  | 3 | 3 | 3 | 3 | 1 |
|  | BLSMOTE | 0.0000 | 0.6905 | 0.0000 | 0.0000 | 0.6905 |
|  |  | 3 | 3 | 3 | 3 | 1 |
|  | SLSMOTE | 0.0000 | 0.6905 | 0.0000 | 0.0000 | 0.6905 |
|  |  | 3 | 3 | 3 | 3 | 2 |
|  | RDS | 0.1512 | 0.7717 | 0.2549 | 0.1941 | 0.5340 |
|  |  | 2 | 2 | 2 | 2 | 1 |
|  | ORIG | 0.0000 | 0.6905 | 0.0000 | 0.0000 | 0.6905 |
|  |  | 3 | 3 | 3 | 3 | 1 |
| MG | SASYNO | 0.3132 | 0.9942 | 0.5573 | 0.4439 | 0.9552 |
|  |  | 4 | 2 | 4 | 4 | 4 |
|  | ADASYN | 0.2536 | 0.9933 | 0.5006 | 0.3746 | 0.9426 |
|  |  | 5 | 3 | 5 | 5 | 5 |
|  | SMOTE | 0.8075 | 0.9877 | 0.8922 | 0.6005 | 0.9852 |
|  |  | 2 | 6 | 2 | 1 | 1 |
|  | BLSMOTE | 0.2474 | 0.9924 | 0.4939 | 0.3615 | 0.9421 |

|  | | 6 | 5 | 6 | 6 | 6 |
|---|---|---|---|---|---|---|
|  | SLSMOTE | 0.5239 | 0.9926 | 0.7185 | 0.5930 | 0.9777 |
|  |  | *3* | *4* | *3* | *2* | *3* |
|  | RDS | 0.1719 | 0.9968 | 0.4121 | 0.2862 | 0.8939 |
|  |  | *7* | *1* | *7* | *7* | *7* |
|  | ORIG | 0.8200 | 0.9857 | 0.8980 | 0.5324 | 0.9838 |
|  |  | *1* | *7* | *1* | *3* | *2* |
| OD | SASYNO | 0.2160 | 1.0000 | 0.4647 | 0.3552 | 0.2372 |
|  |  | *6* | *1* | *6* | *5* | *6* |
|  | ADASYN | 0.2214 | 0.9980 | 0.4700 | 0.3625 | 0.2612 |
|  |  | *5* | *6* | *5* | *4* | *5* |
|  | SMOTE | 0.2513 | 0.9994 | 0.5011 | 0.4016 | 0.3741 |
|  |  | *2* | *2* | *2* | *1* | *2* |
|  | BLSMOTE | 0.2292 | 0.9988 | 0.4784 | 0.3729 | 0.2936 |
|  |  | *3* | *4* | *3* | *2* | *3* |
|  | SLSMOTE | 0.2226 | 0.9982 | 0.4714 | 0.3641 | 0.2666 |
|  |  | *4* | *5* | *4* | *3* | *4* |
|  | RDS | 0.2127 | 0.9991 | 0.4610 | 0.3508 | 0.2222 |
|  |  | *7* | *3* | *7* | *6* | *7* |
|  | ORIG | 1.0000 | 0.7916 | 0.8897 | 0.02030 | 0.7920 |
|  |  | *1* | *7* | *1* | *7* | *1* |
| Average Rank | SASYNO | *4* | *1.2* | *3.8* | *2.4* | *3.2* |
|  | ADASYN | *4* | *3.2* | *4.2* | *3.4* | *2.8* |
|  | SMOTE | *2* | *3.6* | *1.8* | *2.8* | *2* |
|  | BLSMOTE | *3.4* | *4* | *3.2* | *3.8* | *3.4* |
|  | SLSMOTE | *2.8* | *3.6* | *2.6* | *3* | *2.6* |
|  | RDS | *4.8* | *2.4* | *4.6* | *4* | *4.4* |
|  | ORIG | *1.4* | *5.2* | *2.2* | *4.6* | *2.4* |

Table 3. Performance comparison between data sampling approaches using KNN as base learner

| Dataset | Algorithm | *SN* | *SP* | *GM* | *FM* | *Acc* |
|---|---|---|---|---|---|---|
| WI | SASYNO | 0.7536 | 0.9335 | 0.8387 | 0.8209 | 0.8528 |
|  |  | *6* | *1* | *6* | *1* | *2* |
|  | ADASYN | 0.8205 | 0.8722 | 0.8459 | 0.7995 | 0.8538 |
|  |  | *5* | *3* | *5* | *2* | *1* |
|  | SMOTE | 0.9163 | 0.7873 | 0.8494 | 0.6961 | 0.8168 |
|  |  | *2* | *6* | *4* | *6* | *5* |
|  | BLSMOTE | 0.8758 | 0.8308 | 0.8530 | 0.7647 | 0.8438 |
|  |  | *4* | *4* | *1* | *3* | *3* |
|  | SLSMOTE | 0.9106 | 0.7990 | 0.8530 | 0.7187 | 0.8262 |
|  |  | *3* | *5* | *2* | *4* | *4* |
|  | RDS | 0.6107 | 0.8901 | 0.7370 | 0.7143 | 0.7408 |
|  |  | *7* | *2* | *7* | *5* | *7* |
|  | ORIG | 0.9333 | 0.7747 | 0.8503 | 0.6712 | 0.8080 |
|  |  | *1* | *7* | *3* | *7* | *6* |
| SB | SASYNO | 0.7439 | 0.8800 | 0.8090 | 0.7829 | 0.8209 |
|  |  | *4* | *2* | *3* | *1* | *2* |
|  | ADASYN | 0.7021 | 0.8897 | 0.7902 | 0.7694 | 0.8005 |
|  |  | *7* | *1* | *7* | *4* | *7* |
|  | SMOTE | 0.7628 | 0.8547 | 0.8073 | 0.7693 | 0.8180 |
|  |  | *2* | *7* | *4* | *5* | *4* |
|  | BLSMOTE | 0.7378 | 0.8656 | 0.7990 | 0.7687 | 0.8112 |
|  |  | *5* | *4* | *5* | *6* | *5* |
|  | SLSMOTE | 0.7598 | 0.8626 | 0.8095 | 0.7753 | 0.8207 |
|  |  | *3* | *5* | *2* | *2* | *3* |
|  | RDS | 0.7226 | 0.8724 | 0.7938 | 0.7671 | 0.8058 |
|  |  | *6* | *3* | *6* | *7* | *6* |
|  | ORIG | 0.7678 | 0.8573 | 0.8112 | 0.7737 | 0.8216 |
|  |  | *1* | *6* | *1* | *3* | *1* |
| GC | SASYNO | 0.4319 | 0.7862 | 0.5820 | 0.5046 | 0.6305 |
|  |  | *5* | *1* | *4* | *1* | *5* |
|  | ADASYN | 0.4233 | 0.7706 | 0.5700 | 0.4844 | 0.6245 |
|  |  | *6* | *2* | *6* | *2* | *6* |
|  | SMOTE | 0.4589 | 0.7522 | 0.5861 | 0.4447 | 0.6665 |
|  |  | *2* | *6* | *2* | *6* | *2* |
|  | BLSMOTE | 0.4473 | 0.7637 | 0.5828 | 0.4705 | 0.6535 |
|  |  | *4* | *3* | *3* | *3* | *4* |
|  | SLSMOTE | 0.4483 | 0.7583 | 0.5815 | 0.4589 | 0.6565 |
|  |  | *3* | *4* | *5* | *5* | *3* |
|  | RDS | 0.3849 | 0.7578 | 0.5386 | 0.4617 | 0.5830 |
|  |  | *7* | *5* | *7* | *4* | *7* |
|  | ORIG | 0.4621 | 0.7504 | 0.5869 | 0.4386 | 0.6685 |
|  |  | *1* | *7* | *1* | *7* | *1* |
| MG | SASYNO | 0.05540 | 0.9940 | 0.2343 | 0.1039 | 0.6629 |
|  |  | *5* | *2* | *5* | *5* | *6* |
|  | ADASYN | 0.05380 | 0.9922 | 0.2303 | 0.1005 | 0.6725 |
|  |  | *6* | *3* | *6* | *6* | *5* |
|  | SMOTE | 0.6548 | 0.9905 | 0.8043 | 0.6264 | 0.9831 |
|  |  | *2* | *6* | *2* | *3* | *2* |
|  | BLSMOTE | 0.4737 | 0.9920 | 0.6844 | 0.5531 | 0.9747 |
|  |  | *4* | *4* | *4* | *4* | *4* |
|  | SLSMOTE | 0.6314 | 0.9919 | 0.7906 | 0.6450 | 0.9830 |
|  |  | *3* | *5* | *3* | *1* | *3* |
|  | RDS | 0.04860 | 0.9961 | 0.2198 | 0.09230 | 0.5829 |
|  |  | *7* | *1* | *7* | *7* | *7* |
|  | ORIG | 0.6747 | 0.9903 | 0.8164 | 0.6298 | 0.9837 |
|  |  | *1* | *7* | *1* | *2* | *1* |
| OD | SASYNO | 0.8433 | 0.9887 | 0.9131 | 0.8975 | 0.9540 |
|  |  | *1* | *2* | *1* | *2* | *1* |
|  | ADASYN | 0.8148 | 0.9845 | 0.8957 | 0.8748 | 0.9432 |
|  |  | *7* | *3* | *7* | *6* | *6* |
|  | SMOTE | 0.8371 | 0.9702 | 0.9012 | 0.8621 | 0.9404 |
|  |  | *5* | *7* | *6* | *7* | *7* |
|  | BLSMOTE | 0.8413 | 0.9840 | 0.9099 | 0.8888 | 0.9505 |
|  |  | *4* | *4* | *2* | *3* | *3* |
|  | SLSMOTE | 0.8416 | 0.9839 | 0.9099 | 0.8886 | 0.9504 |
|  |  | *3* | *5* | *3* | *4* | *4* |
|  | RDS | 0.8289 | 0.9943 | 0.9078 | 0.8979 | 0.9532 |
|  |  | *6* | *1* | *5* | *1* | *2* |
|  | ORIG | 0.8418 | 0.9835 | 0.9099 | 0.8882 | 0.9503 |
|  |  | *2* | *6* | *4* | *5* | *5* |
| Average Rank | SASYNO | *4.2* | *1.6* | *3.8* | *2* | *3.2* |
|  | ADASYN | *6.2* | *2.4* | *6.2* | *4* | *5* |
|  | SMOTE | *2.6* | *6.4* | *3.6* | *5.4* | *4* |
|  | BLSMOTE | *4.2* | *3.8* | *3* | *3.8* | *3.8* |
|  | SLSMOTE | *3* | *4.8* | *3* | *3.2* | *3.4* |
|  | RDS | *6.6* | *2.4* | *6.4* | *4.8* | *5.8* |
|  | ORIG | *1.2* | *6.6* | *2* | *4.8* | *2.8* |

Table 4. Overall ranks of involved data sampling approaches

| Algorithm | Rank |
|---|---|

|         | SN  | SP  | GM  | FM  | Acc | Overall |
|---------|-----|-----|-----|-----|-----|---------|
| SASYNO  | 4.3 | **1.5** | 3.9 | **2.1** | 3.1 | **3** |
| ADASYN  | 5   | 2.7 | 4.7 | 3.5 | 3.9 | 3.9 |
| SMOTE   | **2.1** | 5.4 | **2.5** | 4.1 | **3** | 3.4 |
| BLSMOTE | 3.8 | 4   | 3.3 | 3.9 | 3.7 | 3.7 |
| SLSMOTE | 3.2 | 3.9 | 2.9 | 3.1 | 3.2 | 3.3 |
| RDS     | 6.1 | 2.7 | 6   | 4.9 | 5.7 | 5.1 |
| ORIG    | 1.7 | 6.2 | 2.8 | 5   | **3** | 3.7 |

Furthermore, we also involve DT, RF and MLP as base learners to evaluate the effectiveness of the proposed SASYNO following the similar experimental protocol used by the previous examples and compare the baseline results obtained with the original data sets. The results are reported in Table 5.

Table 5. Performance of SASYNO with alternative classifiers as base learners

| Dataset | Base Learner | Algorithm | SN | SP | GM | FM | Acc |
|---------|-------------|-----------|--------|--------|--------|--------|--------|
| WI | DT  | SASYNO | 0.7332 | 0.9452 | 0.8325 | 0.8170 | 0.8454 |
|    |     | ORIG   | 0.8917 | 0.7895 | 0.8390 | 0.6971 | 0.8140 |
|    | RF  | SASYNO | 0.7639 | 0.9714 | 0.8614 | 0.8505 | 0.8738 |
|    |     | ORIG   | 0.9472 | 0.7935 | 0.8669 | 0.7137 | 0.8282 |
|    | MLP | SASYNO | 0.7053 | 0.9260 | 0.8080 | 0.7888 | 0.8206 |
|    |     | ORIG   | 0.0875 | 0.6265 | 0.0745 | 0.0072 | 0.6264 |
| SP | DT  | SASYNO | 0.8808 | 0.9394 | 0.9096 | 0.8941 | 0.9159 |
|    |     | ORIG   | 0.8912 | 0.9336 | 0.9121 | 0.8941 | 0.9170 |
|    | RF  | SASYNO | 0.9297 | 0.9451 | 0.9374 | 0.9218 | 0.9392 |
|    |     | ORIG   | 0.9335 | 0.9443 | 0.9389 | 0.9228 | 0.9402 |
|    | MLP | SASYNO | 0.6998 | 0.8483 | 0.7419 | 0.7144 | 0.8124 |
|    |     | ORIG   | 0.7100 | 0.8383 | 0.7371 | 0.6874 | 0.7745 |
| GC | DT  | SASYNO | 0.4971 | 0.7724 | 0.6178 | 0.4918 | 0.6860 |
|    |     | ORIG   | 0.4992 | 0.7763 | 0.6215 | 0.5016 | 0.6880 |
|    | RF  | SASYNO | 0.6161 | 0.7934 | 0.6982 | 0.5510 | 0.7485 |
|    |     | ORIG   | 0.6214 | 0.7854 | 0.6975 | 0.5312 | 0.7470 |
|    | MLP | SASYNO | 0.5224 | 0.8222 | 0.6544 | 0.5748 | 0.7050 |
|    |     | ORIG   | 0.6050 | 0.7670 | 0.6797 | 0.4796 | 0.7315 |
| MG | DT  | SASYNO | 0.3060 | 0.9931 | 0.5496 | 0.4288 | 0.9549 |
|    |     | ORIG   | 0.6310 | 0.9897 | 0.7894 | 0.5961 | 0.9821 |
|    | RF  | SASYNO | 0.3842 | 0.9940 | 0.6171 | 0.5079 | 0.9659 |
|    |     | ORIG   | 0.8499 | 0.9895 | 0.9165 | 0.6724 | 0.9874 |
|    | MLP | SASYNO | 0.2559 | 0.9959 | 0.5035 | 0.3910 | 0.9386 |
|    |     | ORIG   | 0.7951 | 0.9873 | 0.8857 | 0.5854 | 0.9847 |
| OD | DT  | SASYNO | 0.8492 | 0.9677 | 0.9065 | 0.8631 | 0.9418 |
|    |     | ORIG   | 0.8503 | 0.9646 | 0.9057 | 0.8589 | 0.9401 |
|    | RF  | SASYNO | 0.8348 | 0.9707 | 0.8996 | 0.8612 | 0.9392 |
|    |     | ORIG   | 0.8591 | 0.9824 | 0.9187 | 0.8950 | 0.9542 |
|    | MLP | SASYNO | 0.8493 | 0.9967 | 0.9196 | 0.9121 | 0.9595 |
|    |     | ORIG   | 0.8589 | 0.9741 | 0.9128 | 0.8758 | 0.9446 |

### 4.2. Numerical examples on benchmark image sets

In this subsection, numerical examples on popular benchmark image sets are presented to demonstrate that SASYNO is a generic approach and can be used to improve the performance of base learners on various classification problems. The following four mage sets are used for experimental investigation. Details of the four datasets are summarized in Table 6, one can find problem descriptions and example images from [31]–[34].

*1)* UCMerced image set (available from http://weegee.vision.ucmerced.edu/datasets/landuse.html) [31];

*2)* WHU-RS19 image set (available from http://captain.whu.edu.cn/repository.html) [32];

*3)* MNIST image set (available from http://yann.lecun.com/exdb/mnist/) [33], and;

*4)* FashionMNIST image set (available from https://github.com/zalandoresearch/fashion-mnist) [34].

The following three classification algorithms are employed as the base learners:

*1)* SONFIS [25];

*2)* SVM [26], and;

*3)* KNN [27].

Here SONFIS uses Euclidean distance as the distance measure, and the level of granularity is set as 12; SVM uses linear kernel; $k$ is equal to 5 for KNN.

Table 6. Details of benchmark image sets for performance evaluation

| Dataset | | # Images | # Classes | # Images per Class | # Features | Resolution |
|---|---|---|---|---|---|---|
| UCMerced | | 2100 | 21 | 100 | 8192×1 | 256×256 |
| WHU-RS19 | | 950 | 19 | 50 | | 600×600 |
| MNIST | Training set | 60000 | 10 | Approximately 6000 | 784×1 | 28×28 |
| | Testing set | 10000 | | Approximately 1000 | | |
| FashionMNIST | Training set | 60000 | | 6000 | | |
| | Testing set | 10000 | | 10000 | | |

In this paper, we use the same approach as described in [25] to extract a $8192 \times 1$ dimensional feature vector from each image of UCMerced and WHU-RS19 image sets using an ensemble feature descriptor formed by the pretrained pre-trained AlexNet [35] and VGG-VD-16 [36] deep learning neural networks. For MNIST and FashionMNIST image sets, we convert them into 784×1 dimensional feature vectors for classifier training and testing, and further normalize them with the corresponding $L_2$ norm following [25]. During the numerical experiments conducted in this subsection, for UCMerced and WHU-RS19, we use SASYNO to create synthetic feature vectors from the feature vectors of training images from each class to double the size of the training sets. For MNIST and FashionMNIST, we are able to create synthetic images from the training images directly thanks to the simpler structure.

Following the common practice, for UCMerced image set, we randomly select out 50% and 80% images per class for synthetic data generation by SASYNO and classifier training, and use the remaining ones for testing [31]. For WHU-RS19 image set, 40% and 60% images per class are randomly selected out for synthetic data generation and classifier training, and the remaining ones are used for testing [32]. The classification accuracy rates on the testing images by the three classifiers trained with the augmented training sets are reported in Table 7, and the results obtained by the classifiers trained with the original training sets are reported as the base line. Note that the reported results in Table 7 are the average of 30 Monte Carlo experiments.

Table 7. Performance of SASYNO on UCMeced and WHU-RS19 image sets

| Dataset | % Training Samples | SONFIS | | SVM | | KNN | |
|---|---|---|---|---|---|---|---|
| | | SASYNO | ORIG | SASYNO | ORIG | SASYNO | ORIG |
| UCMerced | 50 | 0.9313 | 0.9267 | 0.9440 | 0.9419 | 0.9108 | 0.9005 |
| | 80 | 0.9565 | 0.9529 | 0.9583 | 0.9583 | 0.9377 | 0.9293 |
| WHU-RS19 | 40 | 0.9386 | 0.9370 | 0.9458 | 0.9439 | 0.9282 | 0.9228 |
| | 60 | 0.9500 | 0.9463 | 0.9589 | 0.9570 | 0.9414 | 0.9370 |

In the second numerical example, we randomly select out 10000, 20000, 30000, 40000, 50000 and 60000 images from the training sets of MNIST and FashionMNIST for training. The classification accuracy rates of the three base learners trained on the augmented and original training sets under different experimental settings are reported in Table 8. Examples of original images of the original MNIST and FashionMNIST image sets and the synthetic ones generated by SASYNO are given in Fig. 4 for better illustration.

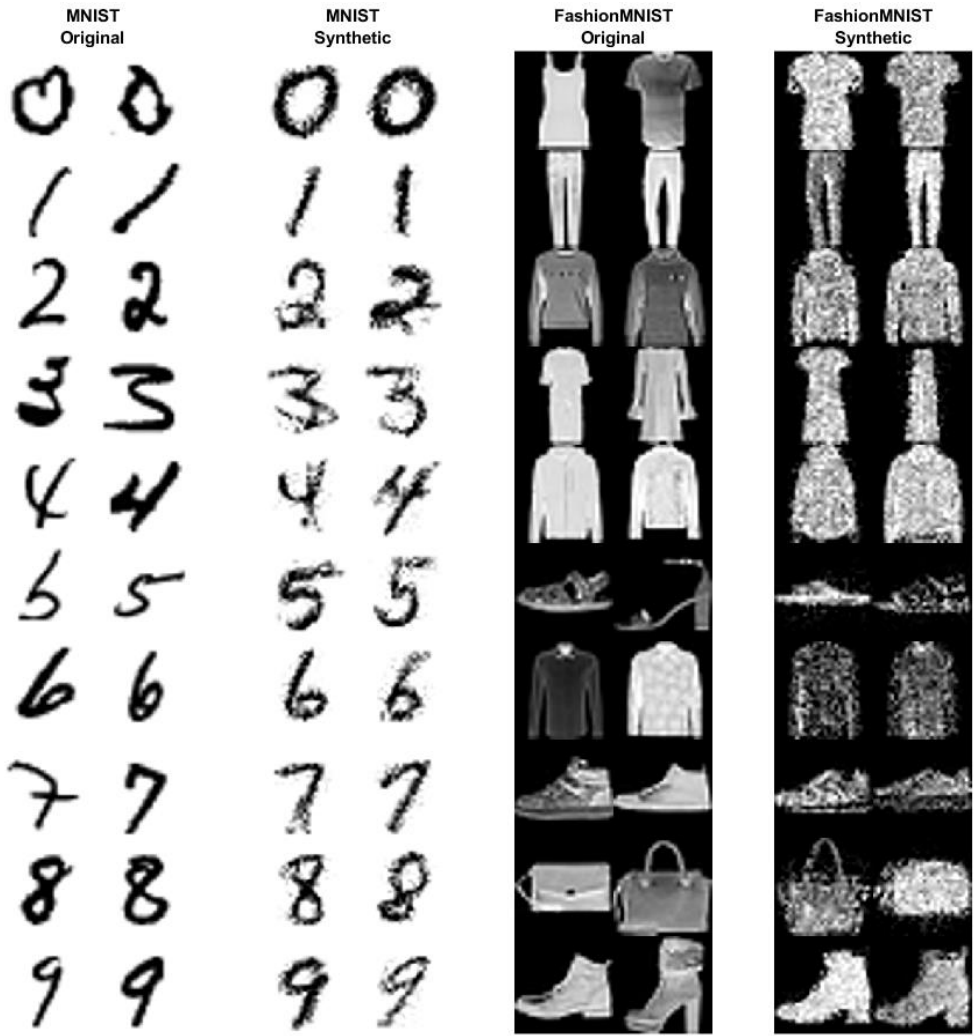

Fig. 4. Examples of original and synthetic images of MNIST and FashionMNIST image sets

Table 8. Performance of SASYNO on MNIST and FashionMNIST image sets

| Dataset | #Training Samples | SONFIS | | SVM | | KNN | |
|---|---|---|---|---|---|---|---|
| | | SASYNO | ORIG | SASYNO | ORIG | SASYNO | ORIG |
| MNIST | 10000 | 0.9512 | 0.9491 | 0.9070 | 0.9037 | 0.9473 | 0.9454 |
| | 20000 | 0.9593 | 0.9588 | 0.9185 | 0.9163 | 0.9571 | 0.9563 |
| | 30000 | 0.9632 | 0.9628 | 0.9232 | 0.9222 | 0.9615 | 0.9614 |
| | 40000 | 0.9659 | 0.9658 | 0.9266 | 0.9261 | 0.9642 | 0.9642 |
| | 50000 | 0.9671 | 0.9682 | 0.9285 | 0.9284 | 0.9665 | 0.9663 |
| | 60000 | 0.9692 | 0.9685 | 0.9306 | 0.9303 | 0.9682 | 0.9679 |
| FashionMNIST | 10000 | 0.8328 | 0.8314 | 0.8314 | 0.8269 | 0.8177 | 0.8164 |
| | 20000 | 0.8450 | 0.8436 | 0.8424 | 0.8387 | 0.8350 | 0.8345 |
| | 30000 | 0.8514 | 0.8529 | 0.8462 | 0.8453 | 0.8445 | 0.8440 |
| | 40000 | 0.8565 | 0.8571 | 0.8490 | 0.8490 | 0.8504 | 0.8499 |
| | 50000 | 0.8600 | 0.8609 | 0.8510 | 0.8508 | 0.8548 | 0.8544 |
| | 60000 | 0.8639 | 0.8633 | 0.8511 | 0.8523 | 0.8582 | 0.8578 |

## 4.3 Discussions

Numerical examples presented in subsection 4.1 demonstrate that the proposed SASYNO can effectively tackle the class imbalance problem by creating high-quality synthetic minority class samples and improve the overall performance of different base learners including SONFIS, KNN, SVM, DT, RF and MLP on various highly imbalanced data sets. Compared with the state-of-the-art data sampling approaches involved in comparison, one can see from Tables 1-4 that SASYNO significantly improves the true negative ratio of the classification results by these base learners (namely, specificity) and outperforms all the comparative data sampling approaches in terms of specificity, F-measure and the overall ranks.

Numerical examples presented in subsection 4.2 further demonstrate the promise of SASYNO as a generic approach for data augmentation, even for very high-dimensional problems. As one can see from Table7, the classification performance of SONFIS, KNN and SVM is improved by involving SASYNO for training set augmentation. On the other hand, as one can see from Table 8, SASYNO effectively improves the classification performance of the three base learners when the scale of the training set is relatively small (10000, 20000 training images). However, the problem of overfitting occurs with the scale of the training set becomes large (30000, 40000, 50000, 60000 training images), and SASYNO is not able to improve the classification performance furthermore.

## 5. Conclusion

This paper presented a new over-sampling approach named SASYNO to tackle the imbalance classification problem. The proposed approach is able to generate high-quality synthetic samples from the empirically observed minority class samples and effective balance the data set. Numerical examples on benchmark binary classification problems demonstrate the better performance of SASYNO comparing with the popular alternatives. In addition, it is justified through numerical examples that SASYNO is a generic approach and can be used for data augmentation for various classification problems.

As future work, we will explore more on imbalanced multi-class classification problems. Such problems are far more challenging for standard classification algorithms compared with binary classification problems. It is important to see how SASYNO perform on these problems. Also, it is shown by numerical examples that SASYNO can be used for creating synthetic images. Lack of labelling is a major problem in field of image recognition, it will be very interesting to see how deep convolutional neural networks react to these synthetic images generated by SASYNO.

## 6. Author Contributions

P. Angelov conceived the original idea, which was developed further by X. Gu. X. Gu designed and implemented the algorithms. X. Gu and E. Almeida Soares designed, performed the experiments and interpreted the results. X. Gu and P. Angelov wrote the manuscript.